\documentclass{styles/svproc}
\usepackage{url}
\usepackage{cite}
\usepackage{amsmath,amssymb,amsfonts}
\usepackage{algorithmic}
\usepackage{graphicx}
\usepackage{textcomp}
\usepackage{xcolor}
\usepackage{float}
\def\BibTeX{{\rm B\kern-.05em{\sc i\kern-.025em b}\kern-.08em
    T\kern-.1667em\lower.7ex\hbox{E}\kern-.125emX}}
\usepackage{lipsum}
\usepackage{graphicx}
\if CLASSOPTION compsoc
\usepackage[caption=false, font=normalsize, labelfont=sf, textfont=sf]{subfig}
\else
\usepackage[caption=false, font=footnotesize]{subfig}
\fi

\usepackage{graphicx}
\usepackage{amssymb}
\usepackage{adjustbox}
\usepackage{colortbl}
\usepackage{comment}
\usepackage{multirow}

\begin{document}
\mainmatter              
\title{Exploring visual language models as a powerful tool in the diagnosis of Ewing Sarcoma}
\titlerunning{Ewing Sarcoma}  
%
\author{Álvaro Pastor-Naranjo$^{1,*}$, Pablo Meseguer$^{1,2}$, Rocío del Amor$^{1}$, Jose Antonio Lopez-Guerrero$^{3}$, Samuel Navarro$^{4,5}$, Katia Scotlandi$^{6}$, Antonio Llombart-Bosch$^{4}$, Isidro Machado$^{4,5,7,8}$, Valery Naranjo$^{1}$ }

\authorrunning{A. Pastor-Naranjo et al.} 
%
\tocauthor{Pablo Meseguer*, Rocío del Amor, Valery Naranjo}
\institute{Instituto Universitario de Investigaci\'{o}n e Innovaci\'{o}n en Tecnolog\'{i}a Centarada en el Ser Humano, HUMAN-tech, Universitat Polit\`{e}cnica de Val\`{e}ncia, Val\`{e}ncia, Spain \\
\and valgrAI - Valencian Graduate School and Research Network of Artificial Intelligence, Val\`{e}ncia, Spain \\
\and Molecular Biology Laboratory, Valencian Oncology Institute, Valencia, España \\
\and Pathology Department, Universitat de Valencia, Valencia, España \\
\and CIBER Cancer (CIBERONC), Madrid, España \\
\and Experimental Pathology Laboratory. Rizzoli Orthopedic Institute, Bologna, Italia \\ 
\and Departamento de Patología, Valencian Oncology Institute, Valencia, España \\
\and Experimental Patologika Laboratory, QuironSalud hospital, Valencia, España \\
\email{\{apasnar,pabmees,madeam2,vnaranjo\}@upv.es} \\ 
}

\maketitle              

\begin{abstract}

Ewing's sarcoma (ES), characterized by a high density of small round blue cells without structural organization, presents a significant health concern, particularly among adolescents aged 10 to 19. Artificial intelligence-based systems for automated analysis of histopathological images are promising to contribute to an accurate diagnosis of ES. In this context, this study explores the feature extraction ability of different pre-training strategies for distinguishing ES from other soft tissue or bone sarcomas with similar morphology in digitized tissue microarrays for the first time, as far as we know. Vision-language supervision (VLS) is compared to fully-supervised ImageNet pre-training within a multiple instance learning paradigm. Our findings indicate a substantial improvement in diagnostic accuracy with the adaption of VLS using an in-domain dataset. Notably, these models not only enhance the accuracy of predicted classes but also drastically reduce the number of trainable parameters and computational costs.

\keywords{Deep learning, tissue microarrays, Ewing sarcoma, visual language models}
\end{abstract}
\section{Introduction}

Sarcomas are neoplasms that arise in the bone or soft tissues and represent more than 20\% of pediatric malignant neoplasms \cite{burningham2012epidemiology}. Specifically, Ewing's sarcoma (ES) is a type of malignant bone tumor characterized by its rapid spread and aggressiveness. Although it is a cancer with a low incidence worldwide, its social impact is considerable, as it mainly affects children and adolescents \cite{paulussen2001ewing}. The 5-year survival rate for ES is highly dependent on the extent of the tumor, decreasing from 68\% in localized tumors to 39\% when metastases are present. These statistics highlight the need for early diagnosis of ES tumors as their prognosis differs from other types of sarcomas, requiring specialized treatment \cite{esiashvili2008changes}.

Histologic analysis of biopsies is a fundamental tool in the diagnosis of ES. The histology of ES predominantly comprises round cells, with smaller proportions of small ovoid and spindle cells arranged in a homogeneous pattern with few stromal tissue. Features are also present in the small round cell tumor (SRCT) family. The histopathologic study can be performed using tissue microarrays (TMA), including multiple circular micrometric sections of cylindrical biopsies (named cores), allowing their comparative analysis. This technique differs from whole biopsy sections that enable detailed observation of the entire tissue. With advances in digital pathology and the advent of specialized scanners, the digitization of histologic tissue samples from biopsies has allowed the implementation of computer vision algorithms based on artificial intelligence (AI).

Previously, deep learning algorithms have been applied to detect Ewing's sarcoma in pediatric radiography  \cite{consalvo2022two}. Regarding histological images, two recent works used machine learning to generate predictive models for the major histological subtypes of rhabdomyosarcoma (a subtype of sarcoma) \cite{frankel2022machine}, \cite{zhang2022deep}, \cite{milewski2023predicting}. In the realm of Ewing Sarcoma research, emphasis primarily revolves around classifications in binary tasks and prognostic evaluations of the cancer\cite{EWING2021Prognosis} rather than its classification into a multiclass problem. With this in mind, our contribution holds significant value as it fills this crucial gap. In concrete, this work proposes a new paradigm based on multiple instance learning (MIL) using histological images as the main source of data to accurately assess the diagnosis of Ewing's sarcoma on tissue microarrays and differentiate it from three other soft tissue or bone sarcomas: chondrosarcoma, gastrointestinal stroma tumor spindle or epithelioid variants and rhabdomyosarcoma. For this purpose, a frozen vision-language model specialized for computational pathology is used to extract high-level feature representations from patches corresponding to the cores. After that, a powerful embedding aggregator based on transformer is used to obtain the prediction at the core level. This methodology is compared with feature extractors that require training, such as the VGG-based architectures, showing the competitiveness of the proposed method with the added value of a highly reduced number of trainable parameters.


\section{Related work}

\subsection{Visual language models}
Contrastive learning has emerged as a powerful pretraining technique
for learning task-agnostic visual features from language supervision. This technique has shown to be a highly effective and scalable strategy to pre-train dual-encoder image-text models that can excel at a range of downstream visual recognition tasks. Representative works such as CLIP \cite{radford2021learning} and ALIGN \cite{jia2021scaling} showed that by scaling to large, diverse web-source datasets of paired images and captions, we can
train models capable of exhibiting fairly robust zero-shot
transfer capabilities through the use of prompts that exploit
the cross-modal alignment between image and text learned
by the model during pretraining. In medical imaging, ConVIRT \cite{zhang2022contrastive} considered paired chest X-ray
images and reports for learning aligned visual language representation. Recent methods CONCH \cite{lu2023towards} and PLIP \cite{huang2023visual} have been used to learn visual representations for
histopathology images using large datasets containing histopathology image-caption pairs from pathology textbooks,
PubMed research articles and Twitter.


\subsection{Multiple instance learning}

In the Multiple Instance Learning (MIL) framework, instances are organized into bags, with labels only assigned at the bag level. According to the standard MIL assumption, a bag is considered positive if it contains at least one instance belonging to the positive class. Among other tasks, it has been used to detect breast cancer \cite{das2020detection} and grade local patterns in prostate cancer \cite{silva2021self}. This assumption makes sense when the labels at the pixel/region label directly affect the image label. However, there are cases where this is not true. In this case, the whole image outcome combines features among the different patches \cite{del2021attention}, called the bag-embedding MIL.
The most common technique is to obtain the bag-level representation by instance-level aggregation of features extracted from each instance by a backbone. The feature extraction is frequently performed with pre-trained networks, transfer learning \cite{zhao2020predicting}, and, more interestingly, following contrastive learning. Then, the aggregation of the patch features results in the bag embedding. The most straightforward and non-trainable aggregation techniques are batch global average (BGAP) \cite{tennakoon2019classification} and batch global max pooling (BGMP) \cite{das2020detection}. Other aggregation techniques include trainable parameters, such as weighted embeddings based on attention \cite{Ilse2018} or recurrent neural networks (RNN) \cite{Campanella2019Clinical}. Recently, authors in \cite{shao2021transmil} proposed a Transformed-based correlated MIL (TransMIL) that considers the morphological and spatial correlation between instances.

\section{Dataset}
\begin{figure*}[t]
\centering
\includegraphics[width=\textwidth]{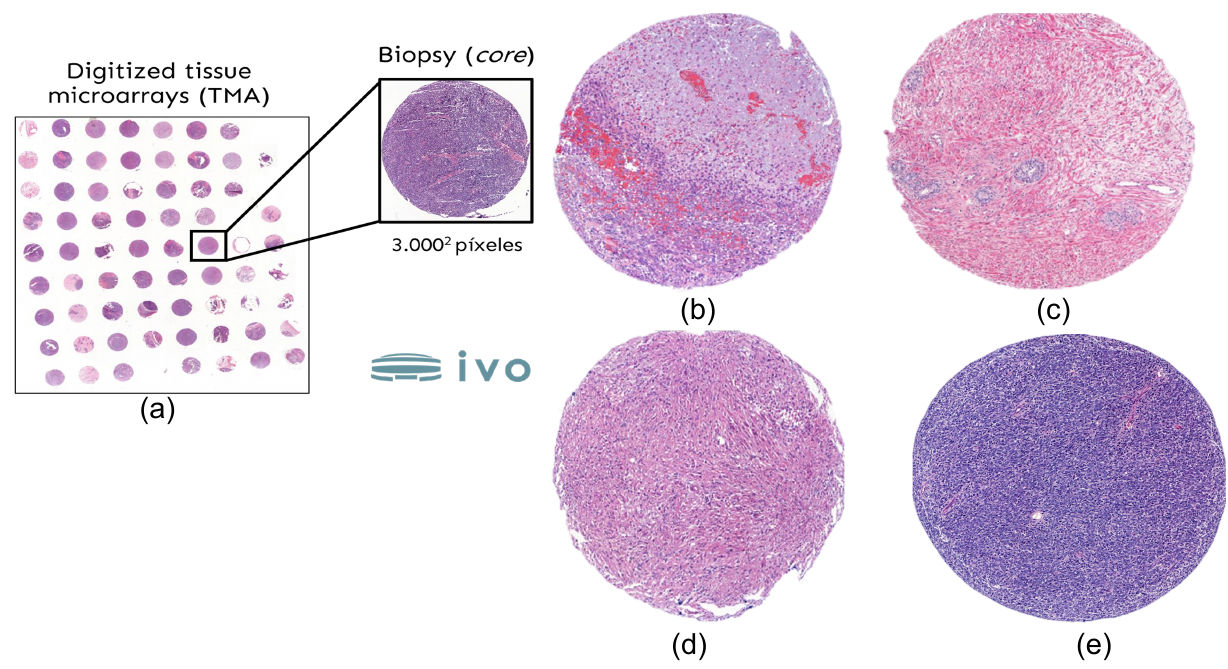}
\caption{(a) Extraction process of cores from digitized microarrays (TMAs); (b) chondrosarcoma (COND), (c) Ewing's sarcoma, (d) gastrointestinal stroma (GIST) and (e) Rhabdomyosarcoma  (RHABDO).}
\label{figure_database}
\end{figure*}

The database used in this study comprises circular micrometric sections of cylindrical biopsies extracted from patients with a genetic confirmation . These samples, followed by histological techniques, were placed in tissue microarrays and digitized with digital pathology scanners to obtain digitized histological images. The database is constituted of different types of sarcomas, including Ewing's sarcoma (EWING), chondrosarcoma (COND), gastrointestinal stroma tumor (GIST) and Rhabdomyosarcoma (RHABDO). Figure \ref{figure_database} shows the histological features of each sarcoma. The database summary, including the number of TMAs and cores, is presented in Table \ref{table_data}. 

\begin{table}[h]
\caption{Database summary.}
\label{table_data}
\centering
\begin{tabular}{|c|c|c|c|c|}
\hline
\multicolumn{1}{|l|}{} & \multicolumn{1}{l|}{\textbf{EWING}} & \multicolumn{1}{l|}{\textbf{COND}} & \multicolumn{1}{l|}{\textbf{GIST}} & \multicolumn{1}{l|}{\textbf{RHABDO}} \\ \hline
\textbf{TMA}           & 26                                  & 6                                  & 4                                                                      & 6                                   \\ \hline
\textbf{CORES}         & 1198                                & 208                                & 159                                                                  & 382                                 \\ \hline
\end{tabular}
\end{table}

Note that some TMA cores were found to be non-informative due to the presence of artifact or necrosis artifacts or other non-tumor tissues. Hence, the number of sections per TMA varies markedly among the different classes. The image size of the sections is about $3000^2$ pixels. To implement weakly supervised learning, cropped patches are extracted from each biopsy with a size of $256^2$ pixels without overlap.

\section{Methods}
\begin{figure*}[h]
\centering
\includegraphics[width=\textwidth]{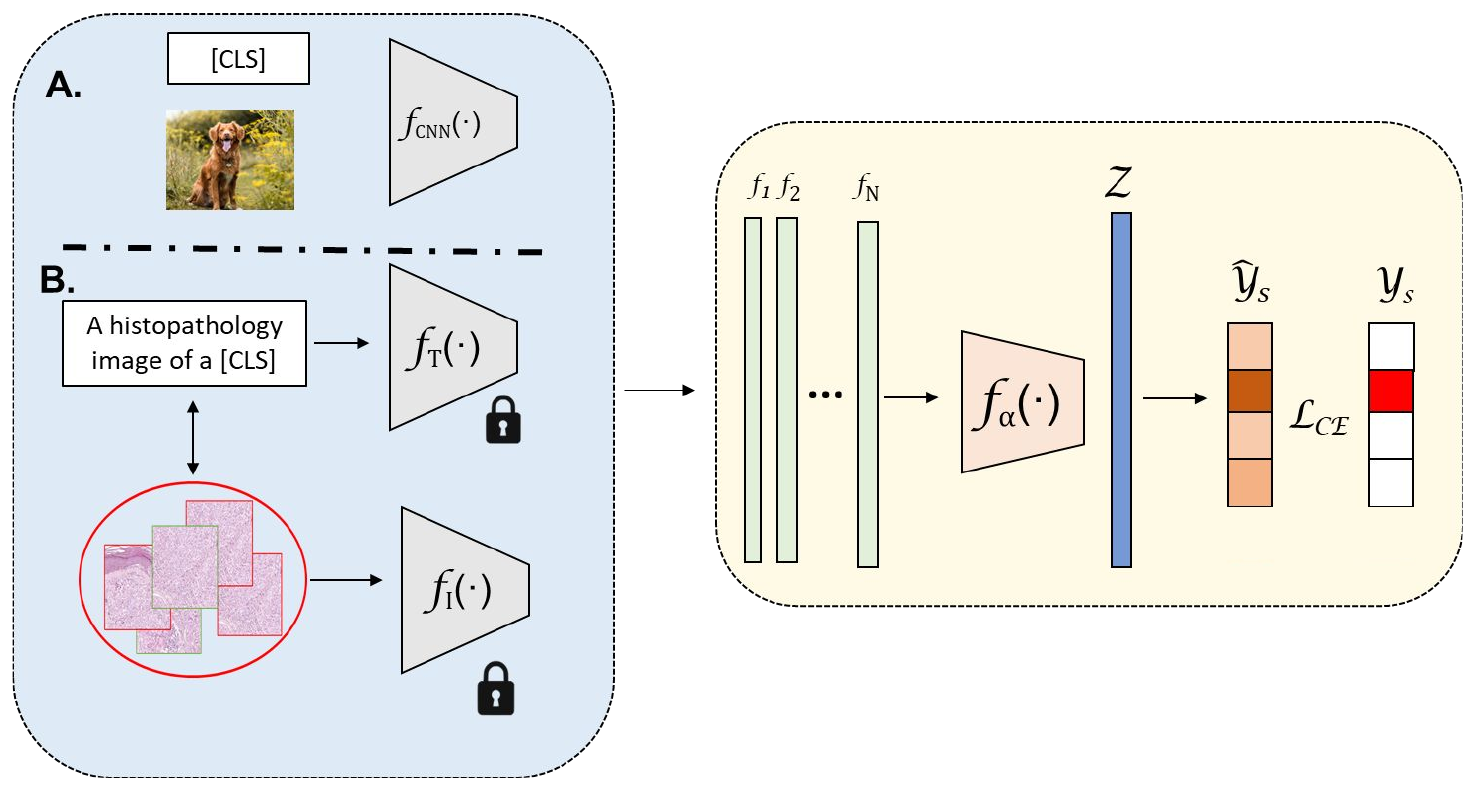}
\caption{Method overview. Multi-class classification of different sarcomas under a multiple instance learning paradigm. In A., feature extraction is described using a CNN-based model, whereas in B., it is based on visual language models. }
\label{fig:summary}
\end{figure*}

An overview of our proposed method is depicted in Figure \ref{fig:summary}. In the following, we describe the problem formulation and each proposed component.

\paragraph{\textbf{Problem Formulation.}} Under the paradigm of MIL, instances are grouped in bags $X = \{x_n\}_{n=1}^{N}$ that exhibit neither dependency nor ordering among them, and its number $N$ is arbitrary for each bag. In the multi-class scenario, each bag is a member of one of $K$ mutually exclusive classes, such that $Y_{k} \in \{0, 1\}$. Note that, in contrast to other MIL formulations, the individual instances do not have an associated label, but rather, the label of the bag is determined by the combination of features of the different instances.

\paragraph{\textbf{Patch-level feature extraction.}}

In the MIL paradigm, the initial step involves patch-level feature extraction. Typically, convolutional neural networks (CNN) architectures, such as VGG16 pre-trained on ImageNet, emerge as the most commonly employed algorithms for this task \cite{del2021attention}. However, the domain in which these networks are pre-trained significantly differs from the histological domain, see Figure \ref{fig:summary} (A.). Fine-tuning such networks for histopathological tasks requires large annotated datasets, which can be challenging to obtain in the medical field due to the time-consuming nature of annotation processes. This underscores the practical limitations of adapting pre-trained networks to histological image analysis tasks within the medical domain. To overcome this limitation, we use Pathology language–image pretraining (PLIP), a multimodal artificial intelligence with both image and text understanding, which is trained on OpenPath, a dataset of paired histopathology image-captions \cite{huang2023visual}. Integrating comprehensive natural language annotations into the learning process increases the capacity to understand image-based semantic knowledge, empowering it to perform a wide range of downstream tasks, see Figure \ref{fig:summary} (B.).

Let us denote a feature extractor, $f_{\theta}(\cdot) : \mathcal{X} \rightarrow \mathcal{Z}$, which projects instances $x \in \mathcal{X}$ to a lower dimensional manifold $z \in \mathcal{F} \subset \mathbb R^{d}$, with $d$ the embedding dimension.

\paragraph{\textbf{Embedding-based MIL}} 
We aim to train a model capable of predicting bag-level labels using a combination of embedding extracted at the instance level. This learning strategy falls under the embedding-based MIL paradigm\footnote{Based on the denomination proposed in \cite{Ilse2018}}.
 We define an aggregation, $f_{\alpha}(\cdot)$, which combines the instance-level projections into a global embedding, $Z$. 
To take into account morphological and
spatial information, we use a transformer-based aggregator \cite{shao2021transmil}. Finally, a neural network classifier, $f_{\phi}(\cdot) : \mathcal{Z} \rightarrow \mathcal{S}$, is in charge of predicting softmax bag-level class scores, $S_{k}$, such that $S_{k} \in [0, 1]$. The optimization of the model parameters is driven by the minimization of standard categorical cross-entropy loss between the reference labels and predicted scores such that:

\begin{equation}
\label{eq:ce}
\mathcal L_{ce} = -\frac{1}{K}\sum_{k=1}^{K} Y_{k} \cdot log(S_{k})
\end{equation}

\section{Results and  Discussion}
\subsection{Implementation details}
The experiments were conducted on the NVIDIA DGX A100 system, with all settings aimed at minimizing cross-entropy during training, incorporating weights that were inversely proportional to the number of instances per class. This approach was aimed at accentuating the impact of misclassifications on minority class images, thereby addressing the challenge of class imbalance. As for the optimizer, AdamW and a learning rate $\eta$ between $1^{-5}$ and $5^{-5}$ were used, depending on the configuration. In terms of batch size, 1 sample per batch was chosen. The database partitioning into training, validation, and test sets was conducted using the following proportions: 60\% for training, 15\% for validation, and 25\% for testing. Model evaluation was carried out employing metrics including sensitivity (SEN), precision (PREC), accuracy (ACC), and the F1-score (F1S).

\subsection{Ablation experiments}
\begin{enumerate} 
\item \textbf{Comparative Analysis of Feature Extractors}
\end{enumerate}

\begin{figure*}[h]
\centering
\includegraphics[width=\textwidth]{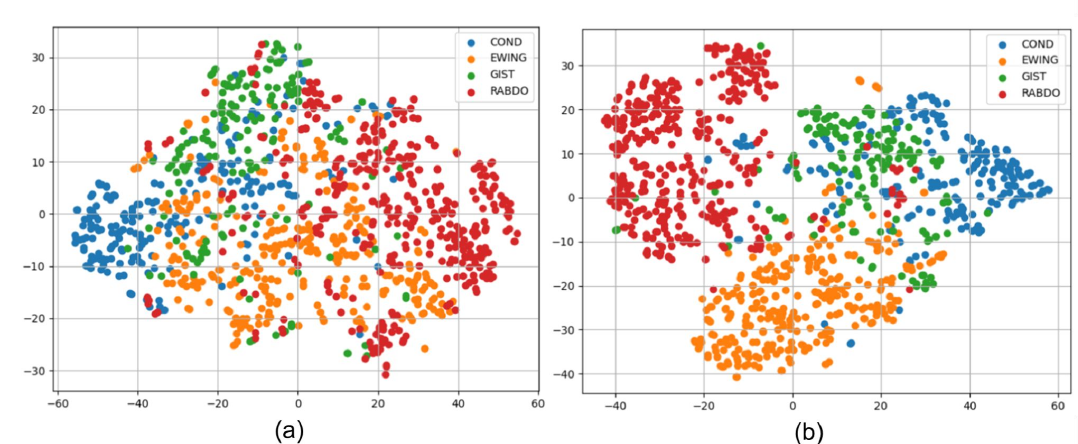}
\caption{TSNE visualization of features embeddings. Core embedding representation from patch features extracted with VGG16 (a) and PLIP (b) for the five neoplasms under study. Note that the core embedding in both cases was calculated by averaging all the patch embedding.}
\label{TSNE}
\end{figure*}

In this section, we present a comparative analysis between PLIP (Pathology Language and Image Pre-Training) and VGG16 regarding their feature representation capabilities. In Figure \ref{TSNE}, it is shown that feature extraction using PLIP yields notably enhanced class separability compared to VGG16.  PLIP's ability to leverage both visual and textual modalities appears to provide a more discriminative representation space, thus facilitating clearer boundaries between different classes. This emphasizes the potential of vision-language models in enhancing feature representation and classification performance in diverse image analysis tasks.

\begin{enumerate} 
\item \textbf{Feature Aggregation Techniques}
\end{enumerate}

\begin{figure*}[h]
\centering
\includegraphics[width=0.7\textwidth]{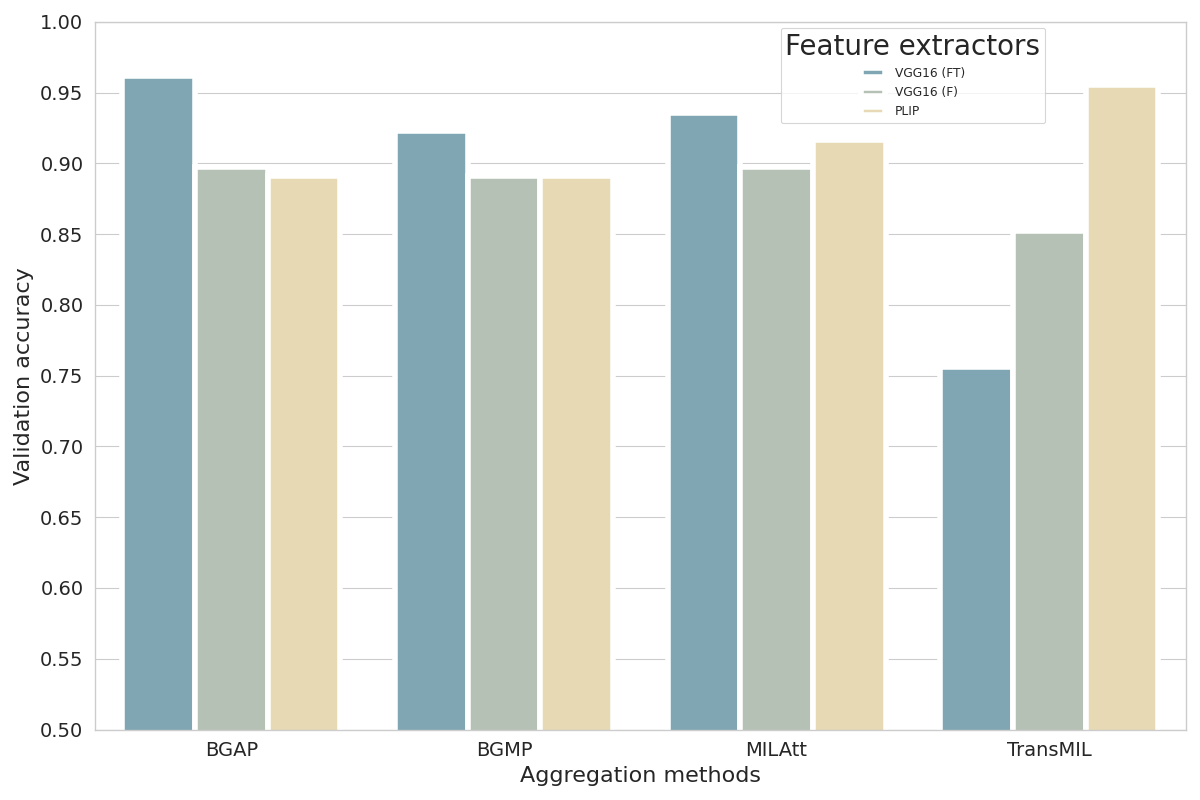}
\caption{Comparison of different embedding aggregators at the core level in the validation set.}
\label{table_validation}
\end{figure*}

We explore various embedding aggregation strategies within the MIL framework: batch global average pooling (BGAP), max pooling (BGMP), attention mechanisms (MILAtt), and a transformer-based aggregator (TransMIL), see Figure \ref{table_validation}. 

Across experiments employing three feature extraction models, fine-tuned VGG (VGG FT), frozen VGG16 (VGG F), and PLIP, the latter consistently demonstrated superior performance when integrated with a transformer-based aggregation method incorporating spatial information. Notably, even with fine-tuning on target domain data, VGG failed to surpass PLIP's efficacy. Our findings underscore a significant advantage: the ability to maintain a frozen feature extractor while focusing training efforts on a robust aggregator. This approach not only enhances computational efficiency but also reduces the time and cost associated with training algorithms on cores.
Considering that the batch size used is one sample and that the most powerful aggregators were used in each case, the fine-tuned VGG takes 0.112 seconds to process the image and update the weights, while using PLIP, this value is reduced to 0.0155 seconds, almost 10 times less. This relation is also true for trainable parameters, 29.7 million for VGG with BGAP versus 2.6 million for PLIP with TransMIL.
This methodological insight contributes to advancing MIL tasks and offers practical implications for resource-efficient model training in histopathological image analysis.

\subsection{Test Results}

\begin{figure*}[h]
\centering
\includegraphics[width=\textwidth]{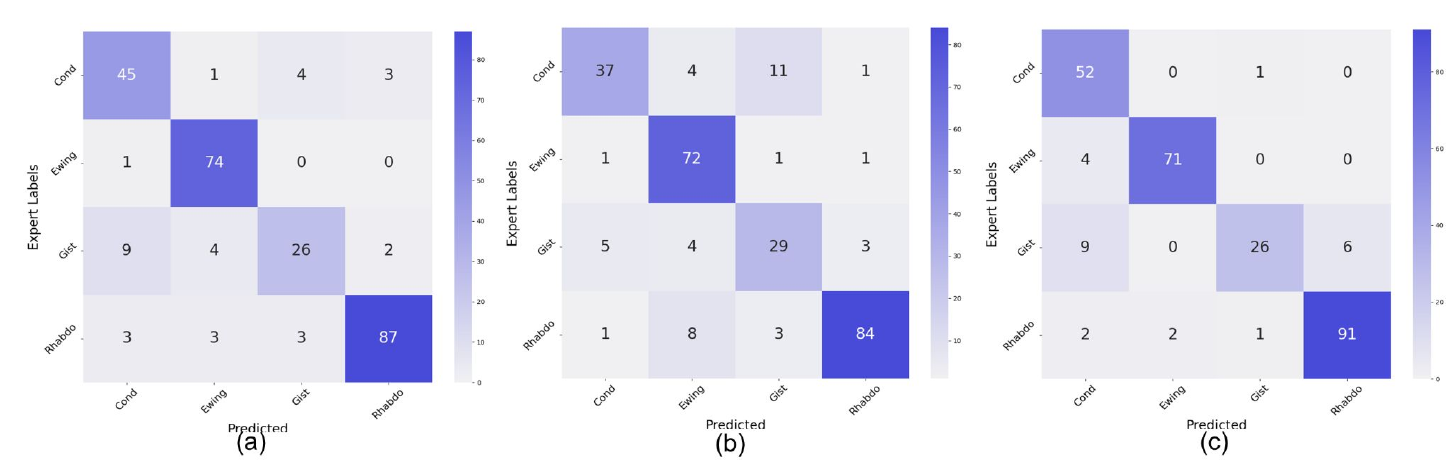}
\caption{Confusion matrix using the test set. (a). VGG16 trained using BGAP as the embedding aggregator. (b) VGG16 frozen using BGAP as embedding aggregation. (c) PLIP frozen using TransMIL as embedding aggregation. }
\label{CM}
\end{figure*}

In the test evaluation, we present the performance results for each feature extractor utilizing their optimal aggregation methods determined during the validation phase: VGG16 (Fine-Tuned) with BGAP, VGG16 with BGAP, and PLIP with TransMIL, see Table \ref{test_results}
and Figure \ref{CM}. 

\begin{table}[h]
\centering
\caption{Test results combining each feature extractor with the best aggregation model obtained in validation.Fine-tunning (FT) and frozen (F).}
\begin{tabular}{|l|c|c|c|c|}
\hline
                    & \multicolumn{1}{l|}{\textbf{SEN}} & \multicolumn{1}{l|}{\textbf{PREC}} & \multicolumn{1}{l|}{\textbf{ACC}} & \multicolumn{1}{l|}{\textbf{F1-S}} \\ \hline
\textbf{VGG16 (FT)} & 0.875                            & 0.875                              & 0.875                             & 0.872                              \\ \hline
\textbf{VGG16 (F)}  & 0.837                             & 0.843                              & 0.837                             & 0.837                              \\ \hline
\textbf{PLIP \cite{huang2023visual}}       & \textbf{0.905}                    & \textbf{0.914}                     & \textbf{0.905}                    & \textbf{0.903}                     \\ \hline
\end{tabular}
\label{test_results}
\end{table}

Notably, our findings align closely with the validation outcomes, affirming the effectiveness of the chosen aggregation techniques. However, a discernible trend emerges, particularly regarding the VGG16 model after fine-tuning. Here, a notable decrease in performance metrics is observed compared to the validation phase, indicating potential overfitting to the training data. This discrepancy underscores the importance of addressing data scarcity and model generalization. 

\section{Conclusion}
In this study, we pioneered the identification of various sarcomas, including Ewing's sarcoma, using histological images through a multiple-instance learning methodology. Furthermore, our investigation highlights the potential of Vision-Language models, which integrate image and textual information during training to improve visual feature learning and efficient model adaptation. We have demonstrated that leveraging such models as frozen patch-level feature extractors and powerful embedding aggregators, such as those based on transformers, surpasses traditional supervised approaches like fine-tuning. Additionally, future lines aim to further advance this field by integrating data augmentation techniques within the MIL framework. 

\section{Fundings}
This work has received funding from the Spanish Ministry of Economy and Competitiveness through projects PID2019-105142RB-C21 (AI4SKIN) and PID2022-140189OB-C21 (ASSIST). The work of Rocío del Amor and Pablo Meseguer has been supported by the Spanish Ministry of Universities under an FPU Grant (FPU20/05263) and valgrAI - Valencian Graduate School and Research Network of Artificial Intelligence, respectively. 

\bibliography{main}
\bibliographystyle{IEEEtran}
%

\end{document}